\documentclass[conference]{IEEEtran}
\IEEEoverridecommandlockouts
\usepackage{cite}
\usepackage{amsmath,amssymb,amsfonts}
\usepackage{algorithmic}
\usepackage{graphicx}
\usepackage{textcomp}
\usepackage{xcolor}
\usepackage{multirow}
\usepackage{booktabs}

\def\BibTeX{{\rm B\kern-.05em{\sc i\kern-.025em b}\kern-.08em
    T\kern-.1667em\lower.7ex\hbox{E}\kern-.125emX}}
\begin{document}

\title{Mamba-Enhanced Text-Audio-Video Alignment Network for Emotion Recognition in Conversations\\

\thanks{This work is partially supported by the National Natural Science Foundation of China (No.62473267), and the Natural Science Foundation of Top Talent of SZTU (No.GDRC202318).}
\thanks{\textsuperscript{*}Xiaomao Fan is the corresponding author.}
}

\author{\IEEEauthorblockN{1\textsuperscript{st} Xinran Li}
\IEEEauthorblockA{\textit{College of Big Data and Internet} \\
\textit{Shenzhen Technology University}\\
Shenzhen, China \\
202100203120@stumail.sztu.edu.cn}
\and
\IEEEauthorblockN{2\textsuperscript{nd} Xiaomao Fan\textsuperscript{*}, Member, IEEE}
\IEEEauthorblockA{\textit{College of Big Data and Internet} \\
\textit{Shenzhen Technology University}\\
Shenzhen, China \\
astrofan2008@gmail.com}
\and
\IEEEauthorblockN{3\textsuperscript{rd} Qingyang Wu}
\IEEEauthorblockA{\textit{College of Big Data and Internet} \\
\textit{Shenzhen Technology University}\\
Shenzhen, China \\
wuqingyang@sztu.edu.cn}
\and
\IEEEauthorblockN{4\textsuperscript{th} Xiaojiang Peng, Senior Member, IEEE}
\IEEEauthorblockA{\textit{College of Big Data and Internet} \\
\textit{Shenzhen Technology University}\\
Shenzhen, China \\
pengxiaojiang@sztu.edu.cn}
\and
\IEEEauthorblockN{5\textsuperscript{th} Ye Li}
\IEEEauthorblockA{\textit{Institute of Advanced Computing and Digital Engineering} \\
\textit{Shenzhen Institutes of Advanced Technology, Chinese Academy of Sciences}\\
Shenzhen, China \\
ye.li@siat.ac.cn}
}

\maketitle

\begin{abstract}
Emotion Recognition in Conversations (ERCs) is a vital area within multimodal interaction research, dedicated to accurately identifying and classifying the emotions expressed by speakers throughout a conversation. Traditional ERC approaches predominantly rely on unimodal cues—such as text, audio, or visual data—leading to limitations in their effectiveness. These methods encounter two significant challenges: 1)\emph{Consistency in multimodal information}. Before integrating various modalities, it is crucial to ensure that the data from different sources is aligned and coherent. 2)\emph{Contextual information capture}. Successfully fusing multimodal features requires a keen understanding of the evolving emotional tone, especially in lengthy dialogues where emotions may shift and develop over time. To address these limitations, we propose a novel Mamba-enhanced Text-Audio-Video alignment network (MaTAV) for the ERC task. MaTAV is with the advantages of aligning unimodal features to ensure consistency across different modalities and handling long input sequences to better capture contextual multimodal information. The extensive experiments on the MELD and IEMOCAP datasets demonstrate that MaTAV significantly outperforms existing state-of-the-art methods on the ERC task with a big margin. The source code is available at URL(https://github.com/Alena-Xinran/MaTAV).
\end{abstract}

\begin{IEEEkeywords}
Emotion recognition in conversations, Multimodal Fusion, Mamba, Emotion classification
\end{IEEEkeywords}

\section{Introduction}
Emotion Recognition in Conversations (ERC) is a critical area of research in the field of multimodal interaction, which focuses on accurately identifying the emotions of speakers throughout various utterances within a conversation. Emotions play a significant role in human communication, influencing decision-making, social interactions, and personal well-being. Hence, developing a reliable ERC system is essential for applications in social media analysis \cite{wang2024multimodal,liu2024emotion}, customer service \cite{chen2024emotion,guo2024measuring}, and mental health monitoring \cite{jiang2024multimodal,fan2023bafnet,ji2023emsn}.

Traditional ERC methods \cite{ezzameli2023emotion,zhu2024emotion} have predominantly relied on unimodal cues such as text, audio, or visual data to classify emotions. While these methods have shown promising results, they often fail to capture the full spectrum of emotional nuances present in conversations. Recent advanced multimodal ERC methods \cite{shi2023multiemo,meng2024deep,yang2024cvan,chen2023multivariate,meng2024masked} have been introduced to address this limitation by integrating information from multiple modalities. However, these multimodal methods still face significant challenges in two main areas: 1)\emph{Achieving consistency across different modalities is challenging because the way emotions are expressed through text can be quite different from how they are conveyed through audio or visual cues}. \emph{e.g.}, "I feel something is a bit off." In the text modality, this phrase might be interpreted as mild concern, while in the audio modality, a trembling tone and rapid breathing could convey a stronger sense of fear, and the tense facial expression in the visual modality further intensifies this emotion. 2)\emph{Effectively capturing the contextual information is crucial for accurate emotion classification}. Existing methods \cite{yang2021hitrans,zhang2021coin} often struggle to incorporate this contextual information during the fusion process, leading to a loss of critical emotional cues. \emph{e.g.}, in the MELD dataset, which features scenes from the TV show Friends, there’s a moment where Ross confesses to Rachel at the airport, saying, "I'm still in love with you." A fixed context window might interpret this as neutral, but by dynamically integrating contextual information can help model accurately captures Ross’s underlying sadness. This is because it considers his prior experiences and fears that this confession won’t change Rachel’s decision.

To address the aforementioned issues, we propose MaTAV, a novel Mamba-enhanced text-audio-video alignment network designed for the ERC task. Inspired by the ALBEF framework \cite{li2021align}, we propose a multimodal emotion contrastive loss (MEC-Loss) to align unimodal features and ensure consistency in multimodal information, alleviating the problem of discrepancy between modalities. Furthermore, we leverage the Mamba network architecture to address the challenge of effectively capturing contextual information during the fusion of multimodal features. Unlike traditional approaches that rely on a fixed context window, the Mamba network dynamically incorporates information from a broader context, making it particularly well-suited for processing long sequences. Additionally, Mamba is optimized for faster inference, allowing it to handle the complexities of long sequences more efficiently. This adaptability and speed make it especially effective in the ERC task, where the emotional tone can evolve over time. We conducted extensive experiments on two widely-used ERC datasets, MELD \cite{poria2018meld} and IEMOCAP \cite{busso2008iemocap}, to evaluate the effectiveness of our proposed MaTAV network. The experiment results demonstrate that MaTAV significantly outperforms the existing state-of-the-art methods in the ERC task. The main contributions of this work can be summarized as follows:

\begin{itemize}
\item[$\bullet$] We first introduce the Mamba network within the ERC task, which significantly enhances the ability to efficiently process lengthy multimodal sequences involving text, audio, and video.
\item[$\bullet$] We propose a novel multimodal emotion contrastive loss, \emph{i.e.} MEC-Loss, designed to tackle the discrepancies between modalities in ERC tasks. The MEC-Loss focuses on aligning unimodal features, ensuring consistency across multimodal information.
\item[$\bullet$] The extensive experimental results on the MELD and IEMOCAP datasets reveal that the MaTAV framework significantly outperforms existing state-of-the-art methods by a considerable margin.
\end{itemize}

\section{Methodology}
\subsection{Overview}
\begin{figure}[tb]
\centering
\includegraphics[width=1.0\linewidth]{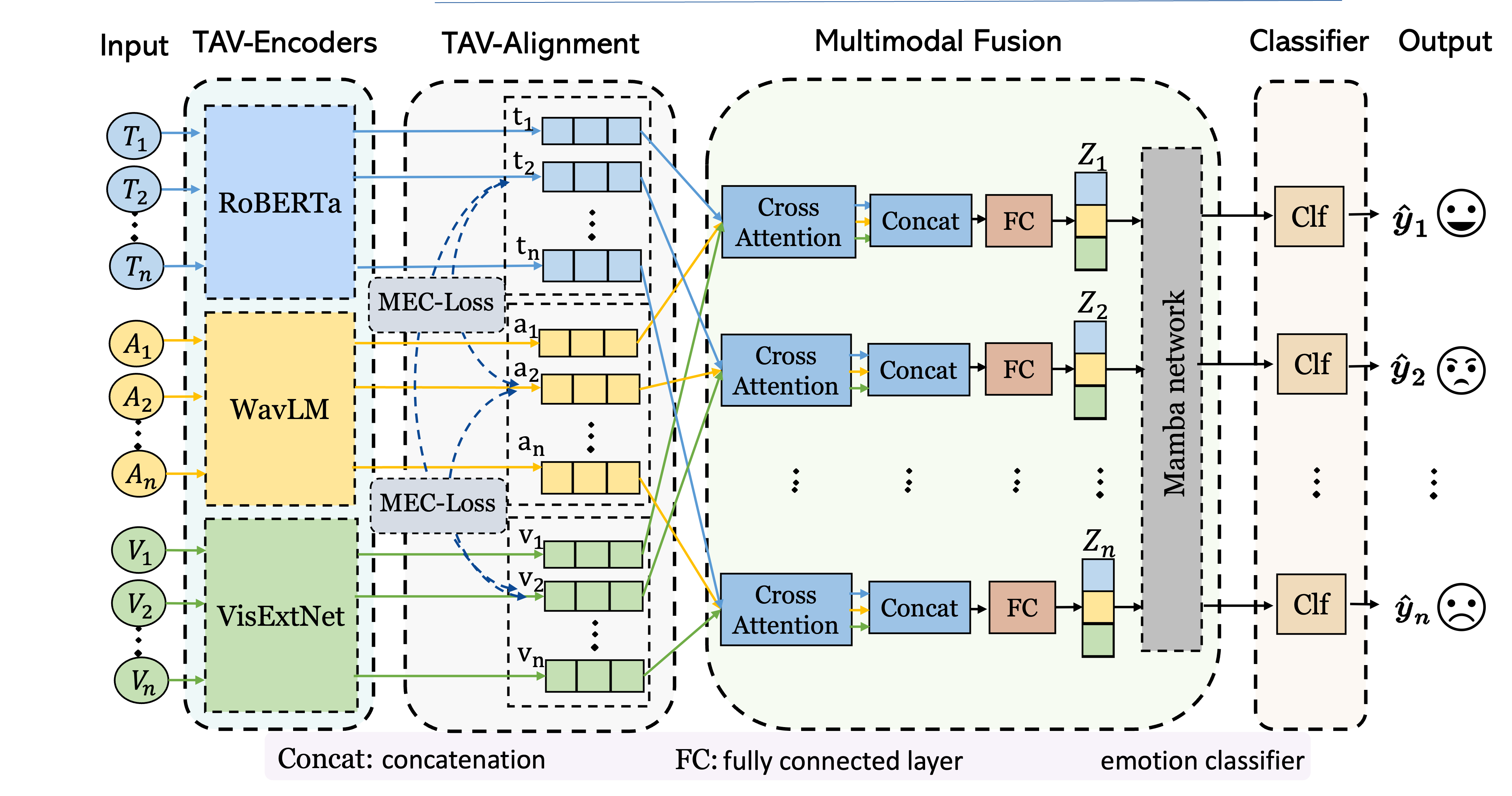}
\caption{The overall network architecture of MaTAV. It consists of four primary components: a text-audio-video encoders (TAV-encoders) module, a text-audio-video alignment (TAV-Alignment) module, a multimodal fusion module, and an emotion classifier.}
\label{model}
\end{figure}

The MaTAV framework, as illustrated in Fig. \ref{model}, consists of four key components: a text-audio-video encoders (TAV-encoders) module, a text-audio-video alignment (TAV-Alignment) module, a multimodal fusion module, and an emotion classifier. The main objective of MaTAV is to accurately identify the emotion label of each utterance from a predefined set of emotion categories. Specifically, the primary input to MaTAV is a dialogue comprising $n$ utterances, which include text ($T$), audio ($A$), and video ($V$) data. For each utterance, specialized encoders—including RoBERTa \cite{kim2021emoberta} for text, WavLM \cite{chen2022wavlm} for audio, and VisExtNet \cite{shi2023multiemo} for video—are employed to extract relevant features from their respective modalities. Building upon the ALBEF approach \cite{li2021align}, we introduce the multimodal emotion contrastive loss (MEC-Loss) to effectively align these extracted features prior to the fusion process.During the multimodal fusion phase, the Mamba network \cite{gu2023mamba} is utilized to capture contextualized information from the integrated multimodal features, facilitating a more nuanced understanding of the emotional context within the dialogue. Finally, an emotion classifier, implemented as a Softmax layer, determines the emotion category associated with each utterance. The following subsections will provide an in-depth exploration of each module within the MaTAV framework, detailing their specific functions and interactions.

\subsection{TAV-Encoders}
In this section, TAV-encoders utilize three specified encoders of RoBERTa \cite{kim2021emoberta}, WavLM \cite{chen2022wavlm}, and VisExtNet \cite{shi2023multiemo} to process input from three modalities of text, audio, and video, respectively. For text encoding, we leverage the RoBERTa which excels in sequence modeling and can be fine-tuned for emotion recognition tasks. RoBERTa transforms the text content \(\{T_1, T_2, \ldots, T_n\}\) into 256-dimensional feature vectors, denoted as \(\{t_1, t_2, \ldots, t_n\}\). In the realm of audio processing, we utilize WavLM as the audio encoder. WavLM is particularly adept at handling adverse conditions, having been trained on extensive unlabeled speech data. The audio content \(\{A_1, A_2, \ldots, A_n\}\) is transformed into 1024-dimensional feature vectors, represented as \(\{a_1, a_2, \ldots, a_n\}\). For video encoding, we adopt VisExtNet, which effectively captures facial expressions while minimizing the inclusion of extraneous visual information. This approach helps to mitigate redundancy in scene-related data. The video content \(\{V_1, V_2, \ldots, V_n\}\)  is processed to yield 1000-dimensional features, denoted as \(\{a_1, a_2, \ldots, a_n\}\).

\subsection{TAV-Alignment}
Inspired by the work of ALBEF \cite{li2021align}, we introduce a novel text-audio-video alignment network called TAV-alignment as shown in Fig.\ref{loss}. This network incorporates a modality contrast loss, referred to as MEC-loss, to facilitate effective alignment among text, audio, and video modalities. By leveraging MEC-loss, TAV-alignment enhances the coherence and synchronization of these diverse data types, improving the ERC performance. Specifically, to compute the MEC-Loss, we first define the similarity scores with cosine between different modalities as follows:

\begin{figure}[tb]
\centering
\includegraphics[width=0.8\linewidth]{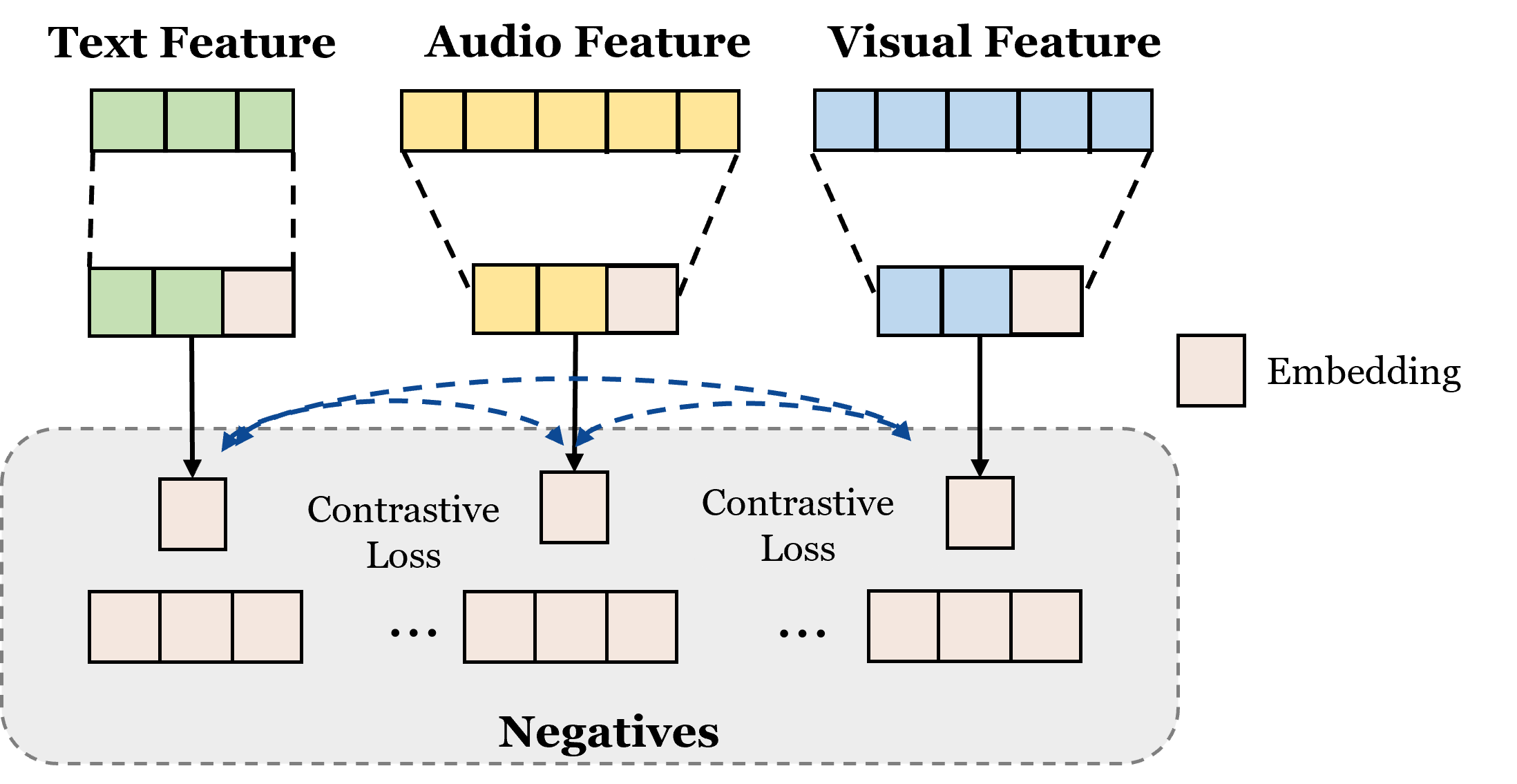}
\caption{The network architecture of Text-Audio-Video Alignment (TAV-Alignment).} 
\label{loss}
\end{figure}

\begin{equation}
    s(q, k) = g_q(q)^\top g_k(k)
\end{equation}

\noindent where $k, q\in\{T, A, V\}$. \(g_q(\cdot)\), and \(g_k(\cdot)\) represent projection heads that map the corresponding embeddings into normalized 256-dimensional representations. Besides, we follow the work of MoCo \cite{he2020momentum}, maintaining three queues to store the most recent \(M\) representations of text, audio, and video from the momentum unimodal encoders, denoted as \(g'_t(t')\), \(g'_a(a')\), and \(g'_v(v')\). The similarity calculations involving momentum encoders are defined as 
\begin{align}
    s(T, A') &= g_t(t)^\top g'_a(a')\\
    s(A, V') &= g_a(a)^\top g'_v(v')\\
    s(T, V') &= g_t(t)^\top g'_v(v') 
\end{align}

 To facilitate effective alignment of the text, audio, and video embeddings in the 256-dimensional space, we present the MEC-Loss aiming to maximize similarity scores for matching pairs while minimizing similarity scores for non-matching pairs, ensuring effective alignment of text, audio, and video embeddings: 

\begin{equation}
\begin{aligned}
\mathcal{L}_{\text{MEC}} = & -\log \frac{\exp(s(T, A'))}{\sum_{j=1}^{M} \exp(s(T, A'_j))} \\
& - \log \frac{\exp(s(A, V'))}{\sum_{j=1}^{M} \exp(s(A, V'_j))} \\
& - \log \frac{\exp(s(T, V'))}{\sum_{j=1}^{M} \exp(s(T, V'_j))}
\end{aligned}
\end{equation}

\subsection{Multimodal Fusion}
Given the limitations of existing methods in the ERC task, we employ the Mamba Network \cite{gu2023mamba} to effectively handle long sequence multimodal data within dialogues. The aligned features of $T$, $A$, and $V$ from the different modalities are fed into a six-layer cross-attention mechanism. The cross-attention is formulated as follows:

\begin{equation} 
    Q = W_Q[t||a||v], K = W_K[t||a||v], V = W_V[t||a||v] 
\end{equation}

\noindent where $W_Q$, $W_K$, and $W_V$ are the weight matrices for the query, key, and value transformations, respectively. The attention computation via the cross-attention mechanism is expressed as:

\begin{equation} 
\text{Attn}(Q, K, V) = \text{Softmax}\left(\frac{QK^\top}{\sqrt{d_k}}\right)V
\end{equation}

\noindent $d_k$ is the dimensional size of $K$. After calculating the attention, the fused features are combined using: $ f = W_F[t||a||v]$, where $W_F$ is the weight matrix for merging the features. Prior to inputting these features into the Mamba Network module, they undergo transformation through a fully connected (FC) layer, yielding outputs denoted as $\{Z_1, Z_2, \ldots, Z_n\}$. To standardize the input dimensions and enhance the model's ability to learn complex relationships within the multimodal data, the features are adjusted to a suitable dimensionality for the Mamba network:$ z_i = \text{FC}(f_i)$. The resulting output features are concatenated into a sequence: $x = [Z_1, Z_2, \ldots, Z_n]$. This sequence is then input into the Mamba network, enabling comprehensive analysis and interpretation of the multimodal data.

\subsection{Emotion Classifier}
The Mamba network is a streamlined end-to-end neural network architecture that operates without attention mechanisms or multi-layer perceptron (MLP) blocks. Following the paradigm of Mamba network, a Softmax layer is employed in the emotion classifier (Clf) to generate a probability distribution across the set of emotion categories. The emotion label with the highest probability is selected as the predicted emotion $\hat{y}_i$ for the 
$i$-th utterance. This process can be mathematically expressed as:
\begin{equation}
\label{eq1}
    \hat{y}_i = \text{argmax}(\text{Softmax}(WF_i + b))
\end{equation}

In the equation (\ref{eq1}), $\hat{y}_i$  represents the predicted emotion label for the $i$-th utterance, while $W$ and $b$ denote the weights and biases of the Softmax layer, respectively. Here, $F_i$ refers to the $i$-th feature in the output sequence $F$. 

\begin{table*}[htbp]
\centering
\caption{Experimental results on the IEMOCAP dataset.}
\label{tab:iemocap}
\begin{tabular}{@{}lccccccc@{}}
\toprule
\textbf{Model} & \textbf{Happiness} & \textbf{Sadness} & \textbf{Neutral} & \textbf{Anger} & \textbf{Excitement} & \textbf{Frustration} & \textbf{$WF1$} \\ \midrule
GA2MIF 2023 \cite{li2023ga2mif}& 46.15 & 84.50 & 68.38 & 70.29 & 75.99 & 66.49 & 70.00 \\
MMGCN 2021 \cite{hu2021mmgcn}& 42.34 & 78.67 & 61.73 & 69.00 & 74.33 & 62.32 & 66.22 \\
LR-GCN 2021 \cite{ren2021lr}& 55.50 & 79.10 & 63.80 & 69.00 & 74.00 & 68.90 & 68.30 \\
DER-GCN 2024 \cite{ai2024gcn}& 58.80 & 79.80 & 61.50 & 72.10 & 73.30 & 67.80 & 69.40 \\
MultiEMO 2023 \cite{shi2023multiemo}& 65.77 & 85.49 & 67.08 & 69.88 & 77.31 & 70.98 & 72.84 \\ \midrule
MaTAV & 63.68 & 85.26 & 74.88 & 70.42 & 75.37 & 70.59 & 73.58 \\
MaTAV$_{w/oMamba}$ & 61.60 & 80.27 & 63.77 & 73.27 & 66.34 & 69.19 & 69.06 \\
MaTAV$_{w/oMEC-Loss}$ & 62.61 & 85.27 & 77.89 & 56.02 & 75.38 & 66.25 & 70.93 \\ \bottomrule
\end{tabular}
\end{table*}

\begin{table*}[htbp]
\centering
\caption{Experimental results on the MELD dataset.}
\label{tab:meld}
\begin{tabular}{@{}lcccccccc@{}}
\toprule
\textbf{Model} & \textbf{Neutral} & \textbf{Surprise} & \textbf{Fear} & \textbf{Sadness} & \textbf{Joy} & \textbf{Disgust} & \textbf{Anger} & \textbf{$WF1$} \\ \midrule
MMGCN 2021 \cite{hu2021mmgcn}& 77.76 & 50.69 & - & 22.93 & 54.78 & - & 47.82 & 58.65 \\
GA2MIF 2023 \cite{li2023ga2mif} & 76.92 & 49.08 & - & 27.18 & 51.87 & - & 48.52 & 58.94 \\
LR-GCN 2021 \cite{ren2021lr}& 80.00 & 55.20 & - & 35.10 & 64.40 & 2.70 & 51.00 & 65.60 \\
DER-GCN 2024 \cite{ai2024gcn} & 80.60 & 51.00 & 10.40 & 41.50 & 64.40 & 10.30 & 57.40 & 66.10 \\
MultiEMO 2023 \cite{shi2023multiemo}& 79.95 & 60.98 & 29.67 & 41.51 & 62.82 & 36.75 & 54.41 & 66.74 \\ \midrule
MaTAV & 82.13 & 55.64 & 25.91 & 38.21 & 64.52 & 30.14 & 56.28 & 66.92 \\
MaTAV$_{w/oMamba}$ & 80.86 & 54.23 & 26.26 & 37.13 & 51.47 & 25.85 & 54.26 & 63.54 \\
MaTAV$_{w/oMEC-Loss}$ & 82.13 & 50.35 & 13.14 & 28.63 & 66.38 & 16.14 & 53.12 & 64.83 \\ \bottomrule
\end{tabular}
\end{table*}

\section{Experimental Settings}
\subsection{Datasets}
In this study, we utilize two public avaliable datasets of IEMOCAP and MELD to evaluate the MaTAV performance.

\noindent \textbf{IEMOCAP} \cite{busso2008iemocap} includes around 12 hours of video recordings of dyadic conversations. These videos are segmented into 7,433 utterances and 151 dialogues. Each utterance is annotated with one of six emotion labels: happiness, sadness, neutral, anger, excitement, and frustration.

\noindent \textbf{MELD} \cite{poria2018meld} is a multi-party dataset derived from the TV series Friends. It consists of 13,708 utterances and 1,433 dialogues. Each utterance is annotated with one of seven emotion categories: anger, disgust, fear, joy, neutral, sadness, and surprise.

\subsection{Implementation Details}
\noindent \textbf{Hyperparameter Settings:} In this paper, the proposed MEMO-Memba is implemented on an Dell-Precision-T7920-Tower workstation using PyTorch 1.8.0, with a Intel(R) Xeon(R) Gold 6248R CPU @ 3.00GHz, 250 GB memory, and an NVIDIA Quadro RTX 6000 GPU with 24 GB VRAM. The parameters for the model training are configured as follows: The batch size is set to 64, and the training will run for 100 epochs. The learning rate is specified at 0.0001, with a weight decay of 0.00001. The model incorporates 6 layers for the cross-attention mechanism. Additionally, the parameter for the MEC loss is set to 0.3.

\noindent \textbf{Evaluation Metrics:} we employ the widely used weighted-average F1 score ($WF1$) as the evaluation metric for the emotion recognition performance of MaTAV.

\section{Results and Discussion}
\subsection{Comparison with Baseline Models}
Compared to existing methods, our MaTAV model achieves state-of-the-art results on both the IEMOCAP and MELD datasets, as illustrated in Tables \ref{tab:iemocap} and \ref{tab:meld}. On the IEMOCAP dataset, MaTAV demonstrates notable advantages across multiple emotion categories, particularly excelling in the Neutral category with the highest $WF1$ score of 74.88\%. The model also shows impressive performance in the Sadness and Frustration categories, achieving $WF1$ scores of 85.26\% and 70.59\%, respectively. On the MELD dataset, MaTAV stands out in the Neutral, Joy, and Anger emotions, with $WF1$ scores of 82.13\%, 64.52\%, and 56.28\%, respectively. It performs well in the Surprise and Sadness categories as well, with scores of 55.64\% and 38.21\%. The superior performance of MaTAV in these categories can be attributed to its advanced dialogue understanding capabilities, which enable it to effectively capture nuanced expressions and variations in emotional tones. This allows for a more accurate and comprehensive emotional analysis in conversational contexts.


\subsection{Ablation Study}
The ablation study focuses on demonstrating the performance contributions of two components of the Mamaba network and MEC-Loss, as shown in Table \ref{tab:iemocap} and Table \ref{tab:meld}. When the Mamba network component is removed, there is a noticeable drop in $WF1$ scores across various emotion categories. Specifically, in the IEMOCAP dataset, the $WF1$ scores for Happiness, Neutral, and Excitement are significantly lower compared to the full MaTAV model. This indicates that the Mamba network component plays a crucial role in accurately capturing these positive emotions. Similarly, the removal of the MEC-Loss component also affects the performance of the model, particularly in the Anger category, where the $WF1$ score drops considerably. This suggests that the MEC-Loss component is essential for the precise classification of more intense emotions such as anger. In the MELD dataset, the absence of the Mamba network component results in lower $WF1$ scores in Surprise, Sadness, and Anger, indicating its importance in handling diverse emotional expressions. The removal of the MEC-Loss component leads to lower scores in Surprise, Fear, and Sadness, underscoring its significance in these negative emotions.

\section{Conclusion}
In this work, we present MaTAV, an innovative multimodal framework specifically designed to enhance performance in the ERC task. By incorporating MEC-Loss for effective alignment of features across text, audio, and video, and utilizing the Mamba Network for dynamic contextual integration, MaTAV adeptly addresses two critical challenges: ensuring consistency in multimodal information and capturing contextual nuances. Experimental evaluations on the MELD and IEMOCAP datasets demonstrate that MaTAV significantly surpasses state-of-the-art methods, resulting in substantial improvements in emotion recognition across various modalities and dialogue turns.

\end{document}